%% file: main.tex
\documentclass[runningheads]{llncs}

\usepackage{pifont}
\usepackage{makecell}
\usepackage{eccv}

\usepackage{eccvabbrv}

\usepackage{graphicx}
\usepackage{booktabs}
\usepackage{amsmath}
\usepackage[accsupp]{axessibility}  % Improves PDF readability for those with disabilities.

\usepackage[pagebackref,breaklinks,colorlinks]{hyperref}
\usepackage{orcidlink}

\begin{document}

% ---------------------------------------------------------------
% TODO REVIEW: Replace with your title
\title{SemanticHuman-HD: High-Resolution Semantic Disentangled 3D Human Generation} 

% TODO REVIEW: If the paper title is too long for the running head, you can set
% an abbreviated paper title here. If not, comment out.
\titlerunning{SemanticHuman-HD}

% TODO FINAL: Replace with your author list. 
% Include the authors' OCRID for the camera-ready version, if at all possible.
\author{Peng Zheng\inst{1} \and
Tao Liu\inst{1} \and
Zili Yi\inst{3} \and
Rui Ma\inst{1,2,}\thanks{Corresponding author}}

% TODO FINAL: Replace with an abbreviated list of authors.
\authorrunning{P.~Zheng et al.}
% First names are abbreviated in the running head.
% If there are more than two authors, 'et al.' is used.

% TODO FINAL: Replace with your institution list.
\institute{{\fontsize{7.0pt}{9.0pt}\selectfont School of Artificial Intelligence, Jilin University, Changchun, China} \and
{\fontsize{7.0pt}{9.0pt}\selectfont Engineering Research Center of Knowledge-Driven Human-Machine Intelligence, MOE, China} \and
{\fontsize{7.0pt}{9.0pt}\selectfont School of Intelligence Science and Technology, Nanjing University, Suzhou, China}
}

\maketitle

\input{paper/abstract}
\input{paper/introduction}

\input{paper/related_work}
\input{paper/method}
\input{paper/experiments}
\input{paper/limitation}
\input{paper/conclusion}

\clearpage  % TODO REVIEW/FINAL: This \clearpage needs to be removed from both review and camera-ready versions.

% ---- Bibliography ----
%
% BibTeX users should specify bibliography style 'splncs04'.
% References will then be sorted and formatted in the correct style.
%
\bibliographystyle{splncs04}
\bibliography{main}
\clearpage 

\appendix
\input{paper/supp}

\end{document}

%% file: paper/abstract.tex
\begin{abstract}
  With the development of neural radiance fields and generative models, numerous methods have been proposed for learning 3D human generation from 2D images. These methods allow control over the pose of the generated 3D human and enable rendering from different viewpoints. However, none of these methods explore semantic disentanglement in human image synthesis, i.e., they can not disentangle the generation of different semantic parts, such as the body, tops, and bottoms. Furthermore, existing methods are limited to synthesize images at $512^2$ resolution due to the high computational cost of neural radiance fields. To address these limitations, we introduce SemanticHuman-HD, the first method to achieve semantic disentangled human image synthesis. Notably, SemanticHuman-HD is also the first method to achieve 3D-aware image synthesis at $1024^2$ resolution, benefiting from our proposed 3D-aware super-resolution module.
  By leveraging the depth maps and semantic masks as guidance for the 3D-aware super-resolution, we significantly reduce the number of sampling points during volume rendering, thereby reducing the computational cost. Our comparative experiments demonstrate the superiority of our method. The effectiveness of each proposed component is also verified through ablation studies. Moreover, our method opens up exciting possibilities for various applications, including 3D garment generation, semantic-aware image synthesis, controllable image synthesis, and out-of-domain image synthesis. 
  \keywords{Generative models \and 3D-aware human image synthesis \and Compositional image synthesis}
\end{abstract}

%% file: paper/introduction.tex
\section{Introduction}
\begin{figure}[t]
    \centering
    \includegraphics[width=\columnwidth]{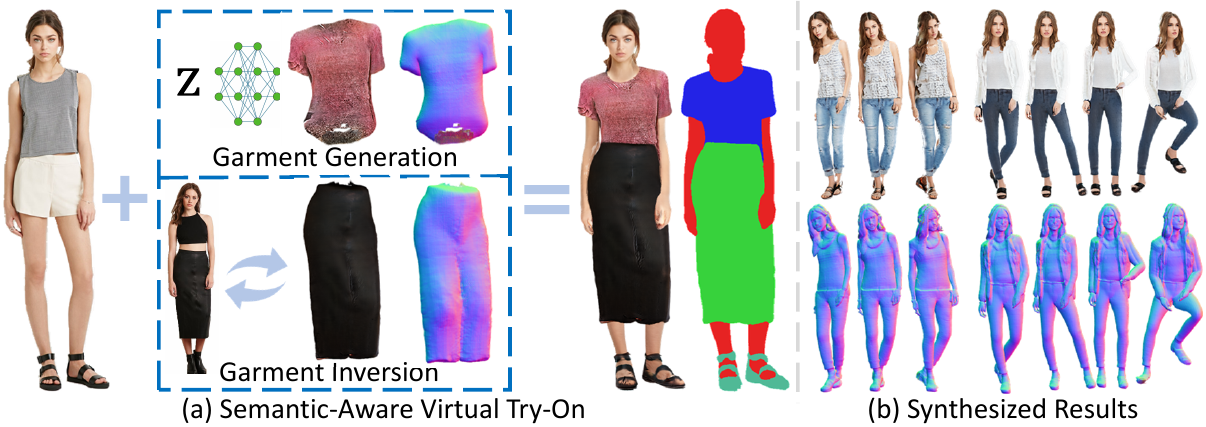}
    \caption{(a) Semantic-aware virtual try-on.  Given a real image, we first employ GAN inversion to obtain its semantic latent code. Subsequently, we replace the top and bottom garment by manipulating the semantic latent code. Here, the top is randomly generated by our model, and the bottom is disentangled from another GAN inversion result. (b) Controllable image synthesis. Our method allows for generating the same person in different poses as well as rendering them from different viewpoints. }
    \label{fig:inv}
\end{figure}
Human image synthesis plays a crucial role in the field of artificial intelligence. This area holds significant potential for applications in virtual reality, virtual try-on, video games, and more. Traditional 2D generative models can only synthesize single-view images. However, recent advancements, such as the development of the neural radiance field (NeRF) \cite{mildenhall2021nerf}, have led to a surge of interest in 3D-aware image synthesis. These methods \cite{or2022stylesdf, xu20223d, chan2022efficient} allow precise control over the viewpoint of synthesized images. While many 3D generative models \cite{or2022stylesdf,chan2022efficient,ma2023semantic,sun2022ide,jiang2022nerffaceediting,sun2022fenerf} focus primarily on portrait synthesis, there is a growing body of work dedicated to full-body human image synthesis. However, none of the existing human image synthesis models fully address semantic disentanglement during generation.  

To achieve semantic disentanglement, some methods \cite{ma2023semantic,zhou2023lc} employ $K$ local 3D generators to model $K$ NeRFs. Each NeRF corresponds to a specific semantic part in the synthesized image. In the case of CNeRF \cite{ma2023semantic}, each generator outputs the color, density and semantic value for a sampled point. The colors outputted by each generator are then weighted and summed, with the weights corresponding to semantic values. While this approach successfully disentangles colors, the geometry of different semantic parts remains entangled. This limitation arises because the densities output by each generator are simply summed. In contrast, 3D-SSGAN \cite{liu20243d} effectively disentangles both texture and geometry. However, it maps 2D feature maps into 3D space, limiting its ability to model complex geometric structures such as full-body humans. Furthermore, the methods mentioned above are specifically designed for portrait synthesis and cannot be naively applied to full-body image synthesis. This limitation arises due to the intricate poses and geometries inherent in the human body. As for full-body image synthesis, AttriHuman-3D \cite{yang2023attrihuman} proposes a framework for semantic-aware human image synthesis, in which decomposed feature planes corresponding to distinct semantic parts are generated using a single 2D generator. While such design of using one generator to generate all semantic parts makes AttriHuman-3D more efficient than previous methods \cite{ma2023semantic,zhou2023lc}, entanglement between different semantic parts is still existed from their results. 

% On the other hand, NeRF \cite{mildenhall2021nerf} suffers from a significant drawback, i.e., high computational cost, since each pixel requires sampling numerous points to accurately integrate colors. Clearly, synthesizing high-resolution images using NeRF-based methods poses challenges. 
On the other hand, synthesizing high-resolution images using NeRF-based methods \cite{mildenhall2021nerf} poses challenges such as high computational cost, each pixel requires sampling numerous points for the accurate integration of colors. 
Some works \cite{dong2023ag3d,zhang2022avatargen,liu20243d,ma2023semantic} employ a super-resolution module to circumvent direct rendering of high-resolution images. However, this strategy might impact the 3D consistency. 
Other works \cite{abdal2023gaussian,hong2022eva3d,chen2023veri3d} propose more efficient way to render high-resolution ($512^2$) images without a super-resolution module. Nevertheless, this resolution may still not satisfy the demands of users, e.g., the need of $1024^2$ images.

To address these issues, we propose SemanticHuman-HD, a novel method for high-resolution human image synthesis with semantic disentanglement. Unlike previous methods, our method generates each semantic part in a completely independent way. 
% In our model, the semantic values introduced earlier are not learned by the models directly; instead, they are normalized from the densities of different NeRFs. The capability to synthesize high-resolution images is provided by our proposed 3D-aware super-resolution module. 
Specifically, we propose a two-stage training process. In the first stage, we synthesize human images, depth maps, semantic masks, and normal maps at $256^2$ resolution. In the second stage, we employ a novel 3D-aware super-resolution module to synthesize $1024^2$ resolution images. This module leverages the depth map and semantic mask synthesized in the first stage as guidance, significantly reducing the computational cost in volume rendering. To demonstrate the superiority of our method, we conduct quantitative and qualitative comparison experiments with state-of-the-art (SOTA) methods. Meanwhile, the effectiveness of each component proposed in this paper is verified in the ablation studies. In summary, our contributions are as follows:
\begin{enumerate}
    \item We propose SemanticHuman-HD, the first method to achieve semantic disentanglement in 3D-aware human image synthesis. In our method, the underlying representation of each part is independent from other parts, leading to exciting applications such as 3D garment generation, semantic-aware virtual try-on, garment-level image editing and out-of-domain image synthesis. 
    % \item In our generation framework, we disentangle both geometry and texture, enabling independent generation of each semantic part. As a result, our method achieves high quality garment generation and . 
    \item Leveraging our 3D-aware super-resolution module, SemanticHuman-HD attains $1024^2$ resolution image synthesis. Importantly, our proposed super-resolution module preserves 3D consistency throughout the synthesis.
    \item Comparing to SOTA human image synthesis methods, our SemanticHuman-HD demonstrates clear superiority in both quantitative measures (e.g., FID) and qualitative evaluation.
\end{enumerate}

%% file: paper/related_work.tex
\section{Related Work}
\subsection{3D-Aware Image Synthesis}
Generative adversarial networks (GANs) \cite{karras2019style,karras2020analyzing,karras2021alias} have demonstrated impressive results on image synthesis tasks. While certain GAN-based methods \cite{tewari2020stylerig, sarkar2021style} achieve pose control in image synthesis, they suffer from a lack of 3D-consistency due to their reliance on 2D feature representations. The advent of neural radiance fields (NeRF) \cite{mildenhall2021nerf,wang2021neus,yariv2021volume,oechsle2021unisurf,barron2021mip} have opened the door to learn 3D-aware image synthesis from 2D image datasets. Numerous works \cite{schwarz2020graf,xu20223d,chan2022efficient,he2023orthoplanes,or2022stylesdf,xu2021generative,chan2021pi,gao2022get3d,gu2021stylenerf} combine NeRF with GAN to achieve 3D-aware image synthesis. Notably, EG3D \cite{chan2022efficient} proposed a tri-plane representation as an efficient alternative to the computationally expensive point-based MLP. Beyond GAN-based generative models, diffusion models have gained prominence in recent years. Several diffusion model-based 3D-aware image synthesis methods  \cite{metzer2023latent,lin2023magic3d,jain2022zero,xu2023dream3d,poole2022dreamfusion,wang2024prolificdreamer,chen2023fantasia3d} have emerged. However, these methods either are general models or only focus on portrait synthesis, making them incapable of synthesize high quality human images. 

\subsection{3D-Aware Human Image Synthesis}
3D-aware human image synthesis faces significant challenges, primarily because humans exhibit articulation and appear in diverse poses and clothing. gDNA \cite{xu2022gdna} introduces a multi-subject forward skinning module for 3D human generation supervised by human scans. Some works \cite{zhang2022avatargen, jiang2023humangen, yang20223dhumangan,grigorev2021stylepeople,zhang2023getavatar,fu2023text,xu2023xagen} leverage human prior \cite{bogo2016keep} and NeRF \cite{mildenhall2021nerf} to learn 3D-aware human image synthesis from 2D image datasets. AG3D \cite{dong2023ag3d} proposes to model the deformation of loose clothing using Fast-SNARF \cite{chen2023fast}. Additionally, it introduces a normal discriminator to improve geometric details in the generated results. EVA3D \cite{hong2022eva3d} adopts a compositional human NeRF representation for high-resolution ($512^2$) 3D-aware human image synthesis, all without relying on super-resolution modules. By leveraging vertex-based radiance fields, VeRi3D \cite{chen2023veri3d} allows local editing of generated results by replacing features at specified vertices. GSM \cite{abdal2023gaussian} is a efficient framework for 3D human generation, which employs Gaussian shell maps to model feature volumes. Similar to VeRi3D, GSM also achieve local editing by a similar way. While some methods \cite{kolotouros2024dreamhuman,hong2022avatarclip,wang2023disentangled,jiang2023mvhuman} focus on 3D human generation using diffusion models, they struggle to synthesize photorealistic images due to limitations inherent in the diffusion model. Notably, none of these methods explore semantic-aware image synthesis. i.e., they cannot edit specific semantic parts of synthesized images while keeping other regions unchanged.

\subsection{Semantic-Aware Image Synthesis}
Some methods \cite{jiang2022nerffaceediting,chen2022sem2nerf,li2021semantic,liu20243d,ma2023semantic,sun2022ide,shi2022semanticstylegan,zhou2023lc,sun2022fenerf,deng20233d} explore semantic-aware image synthesis, supervised by semantic masks. To translate a single-view semantic mask into a NeRF, Sem2NeRF \cite{chen2022sem2nerf} encodes the mask into latent code, controlling the 3D scene representation of a pre-trained decoder. Unlike Sem2NeRF, which requires a semantic mask as input, NeRFaceEditing \cite{jiang2022nerffaceediting} and IDE-3D \cite{sun2022ide} aim to achieve 3D-aware paired semantic mask and image synthesis by learning a semantic mask volume. Specifically for human image synthesis, 3D-SGAN \cite{zhang20223d} proposes a semantic-guided architecture comprising two generators: one for 3D-aware semantic mask synthesis and the other for translating the semantic mask into the corresponding image. Nevertheless, in these methods, different semantic parts are entangled during synthesis. 

Several methods \cite{ma2023semantic,shi2022semanticstylegan,liu20243d} have explored semantic disentangled synthesis. SemanticStyleGAN \cite{shi2022semanticstylegan} uses $K$ local generators to generate $K$ semantic parts in synthesized image. These generators are supervised by paired portraits and semantic masks. The design of $K$ local generators ensures semantic disentanglement, enabling shape and texture changes in specific semantic regions while preserving others. CNeRF \cite{ma2023semantic} and LC-NeRF \cite{zhou2023lc} extend SemanticStyleGAN into the realm of 3D-aware image synthesis by learning compositional NeRFs. 3D-SSGAN \cite{liu20243d} lifts the 2D generator into 3D space for efficiency and stronger disentanglement. Unfortunately, the aforementioned semantic disentangled methods can only be used in portrait synthesis due to the complexities of human poses. 

For human image synthesis, both VeRi3D \cite{chen2023veri3d} and \cite{abdal2023gaussian} can achieve coarse-grain disentanglement during inference, as their features are closely tied to SMPL \cite{bogo2016keep} vertexes, although these vertexes do not align with semantic masks. On the other hand, AttriHuman-3D \cite{yang2023attrihuman} achieves semantic-aware human image synthesis, but it relies on a single generator to produce all semantic parts, lacking true semantic disentanglement. In summary, no existing method achieves 3D-aware human image synthesis with full semantic disentanglement. Furthermore, all these methods, whether utilizing the super-resolution module or not, can only synthesize $512^2$ resolution images. In contrast, our proposed method achieve 3D-aware human image synthesis by independently generating each semantic part. Additionally, we introduce a 3D-aware super-resolution module, enabling the synthesis of $1024^2$ resolution images. 

%% file: paper/method.tex
\section{Method}
\begin{figure}[t]
    \centering
    \includegraphics[width=\textwidth]{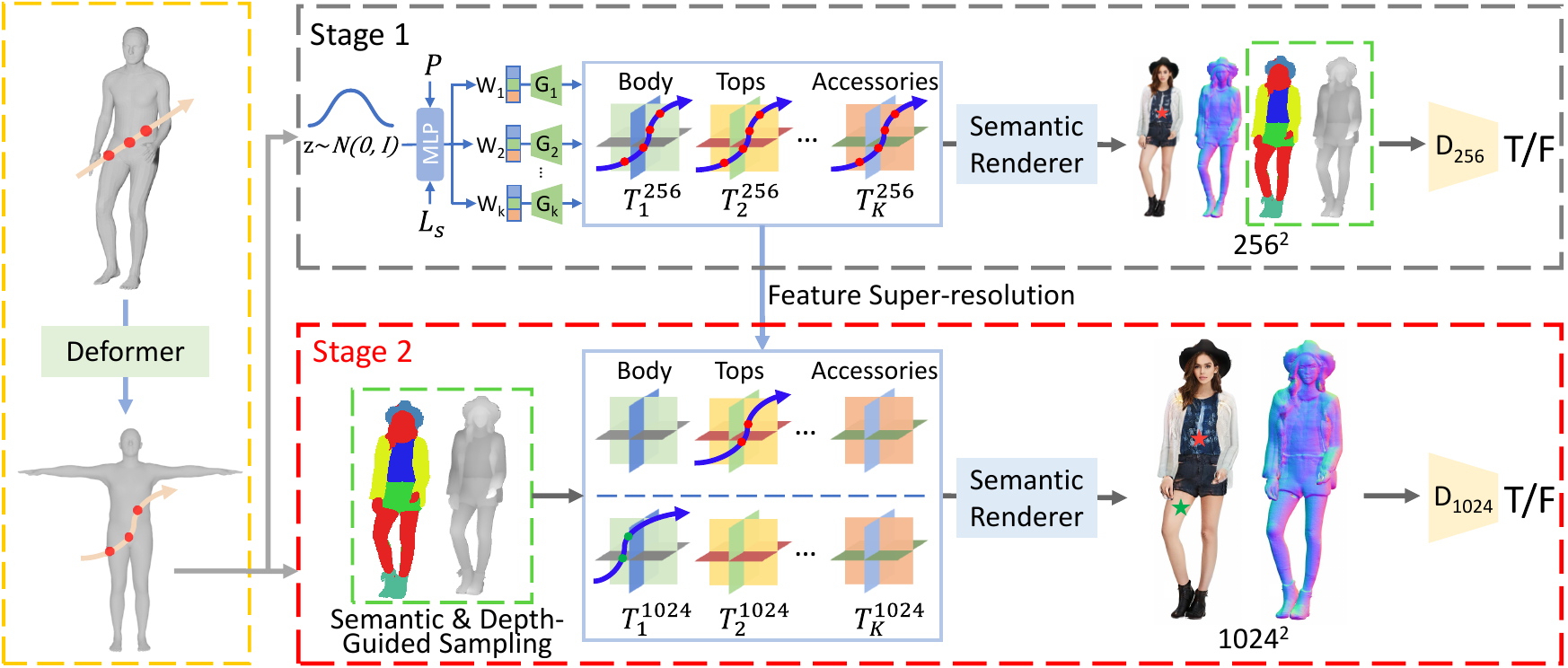}
    \caption{Pipeline of SemanticHuman-HD. In stage 1, given random noise $z$, the Semantic Mapper maps it to $K$ latent code $W_k$, conditioned on human pose $P$ and semantic label $L_s$. Each local generator $G_k$ then maps $W_k$ into a tri-plane representation $T_k^{256}$. For each pixel in the synthesized image, we sample 72 points in posed space, which are subsequently deformed into canonical space using the deformer. These sampled points allow us to interpolate within the tri-plane representation, obtaining color and density information for each point. Finally, the Semantic Renderer renders the image, depth map, semantic mask, and normal map at $256^2$ resolution. In Stage 2, we employ a convolutional network to obtain high-resolution tri-plane representations, denoted as $T_k^{1024}$. To enhance efficiency, we significantly reduce the number of sampling points per pixel using semantic and depth-guided sampling. Ultimately, we render the image and normal map at $1024^2$ resolution. }
    \label{fig:pipeline}
\end{figure}

The overview of our method is depicted in Fig. \ref{fig:pipeline}, and it comprises two stages. In the first stage, given the human pose $P$ and semantic label $L_s$, each generator $G_k$ generates a tri-plane representation, which models one semantic part. The $K$ tri-plane representations are further rendered into a $256^2$ resolution image, depth map, semantic mask and normal map using the semantic renderer. Moving to the second stage, we feed the $K$ tri-plane representations into the 3D-aware super-resolution module. This module enhances the resolution of each tri-plane representation, resulting in higher-quality outputs. Specifically, these refined tri-plane representations can be rendered into a $1024^2$ resolution image, with the depth map and semantic mask serving as guidance. The details of each component are introduced below. 

\subsection{Semantic Disentangled Neural Radiance Field}
\subsubsection{Semantic Mapper:} Given random noise $z$ sampled from a Gaussian distribution, the Semantic Mapper maps it to latent code $W$, conditioned on human pose $P$ and semantic label $L_s$. Here, the pose $P$ corresponds to the parameter of SMPL \cite{bogo2016keep}, while semantic label $L_s$ indicates whether the image contains specific types of garments, e.g., dress, skirt or hat. The latent code $W$ is further extended into a semantic latent code $W^+$, where $W^+ = W^1\times W^2...\times W^K$ and each $W^k$ controls the generation of the $k_{th}$ semantic part. In theory, $K$ can take any value, but in our method, we set $K$ to 6, corresponding to the body, tops, outer, bottoms, shoes and accessories, respectively. During training, we enforce $W^1=W^2...= W^K$ to ensure that different local generators generate consistent parts. For example, men typically do not wear skirts or dresses. During inference, simply modifying $W^k$ allows us to edit the synthesized image in the $k_{th}$ semantic part. 

\subsubsection{Local Generator:} Similar to CNeRF \cite{ma2023semantic}, we employ $K$ local 3D generators to model $K$ semantic parts. However, unlike CNeRF, the generation of different semantic parts in our method are entirely independent. This independence is the key idea behind disentangling both geometry and texture. Conditioned on pose $P$ and semantic label $L_s$, each local generator $G_k$ maps the latent code $W_k$ into a tri-plane representation. For each sampled point $x$, we calculate the density $\sigma(x)$, color $c(x)$, normal $n(x)$ and semantic value $s(x)$ as follows:
\begin{equation}
    \sigma(x) = \sum_{k=1}^K (\sigma(x)^k),\ c(x) = \sum_{k=1}^K c(x)^k\times s(x)^k,
\end{equation}

\begin{equation}
    s(x) = {\rm Concatenate}(s(x)^1, s(x)^2,...,s(x)^K),
\end{equation}

\begin{equation}
    n(x)= \nabla_x(d(x)^{SMPL} + \sum_{k=1}^K \Delta d(x)^k),
\end{equation}

\begin{equation}
\label{eq:sigmoid}
    s(x)^k = \frac{\sigma(x)^k}{\sum_{k=1}^K \sigma(x)^k},\ \sigma(x)^k = Sigmoid(d(x)^{SMPL}+\Delta d(x)^k).
\end{equation}

Here, $\Delta d(x)^k$ and $c(x)^k$ are sampled from $k_{th}$ tri-plane representation, while $d(x)^{SMPL}$ is sampled from the signed distance field (SDF) of canonical SMPL model \cite{bogo2016keep}. Notably, as demonstrated in Eq. \ref{eq:sigmoid}, we initially convert local SDF $\Delta d(x)^k$ to local density $\sigma (x)^k$ and then sum them to obtain the final density $\sigma (x)$. This approach differs from \cite{yang2023attrihuman,zhou2023lc,ma2023semantic}, which first sum local SDF $\Delta d(x)^k$ to obtain final SDF $d(x)$ and subsequently convert it to density $\sigma(x)$. This difference allows us to obtain local density $\sigma(x)^k$ and then map it into semantic value $s(x)^k$, thereby enabling the disentanglement of geometry across different semantic parts. For further details on this distinction, please ref to \cite{ma2023semantic, yang2023attrihuman,zhou2023lc}. 

\subsubsection{Semantic Renderer:} Similar to NeRF \cite{mildenhall2021nerf}, we cast a ray ${\rm r}$ for each pixel along its view direction $v$ from camera center $o$: ${\rm r}(t) = {\rm o} + t{\rm v}$. The color ${\rm C(r)}$, semantic mask ${\rm S(r)}$, depth ${\rm D(r)}$ and normal ${\rm N(r)}$ of each ray $r$ can be rendered as follows:
\begin{equation}
\label{eq:color}
    {\rm \Phi(r)} = \int_{t_n}^{t_f}T(t)\cdot \sigma({\rm r}(t))\cdot \phi({\rm r}(t))dt
\end{equation}

\begin{equation}
\label{eq:depth}
    {\rm D(r)} = \int_{t_n}^{t_f}T(t)\cdot \sigma({\rm r}(t))\cdot tdt,
\end{equation}

\begin{equation}
    {\rm where }\ T(t) = {\rm exp}(-\int_{t_n}^t\sigma ({\rm r}(s))ds).
\end{equation}
In the equations above, $(\Phi,\ \phi)$ represents universal symbols,that can correspond to $({\rm C},\ c)$, $({\rm S},\ s)$ or $({\rm N},\ n)$. For clarity, we intentionally omitted details about the deformer, so in Eq. \ref{eq:color} and Eq. \ref{eq:depth}, each point ${\rm r}(t))$ is actually transformed from posed space to canonical space. For further information about the deformer, please ref to AG3D \cite{dong2023ag3d}.

\subsection{3D-Aware Super-Resolution Module}
In stage 2, we train a 3D-aware super-resolution module to synthesize $1024^2$ resolution images, building upon the $K$ local generator pre-trained in stage 1. The core concept behind this module lies in leveraging semantic mask and depth map synthesized during stage 1 to significantly reduce the number of sampling points. Theoretically, our method achieves a remarkable reduction, i.e., from 432 to 11, where $432=72\times6$. Here, 72 represents 36 points for uniform sampling and 36 points for importance sampling, while 6 corresponds to the number of local generators. Even in the general case, without relying on our specialized semantic disentangled generation, our proposed module can still reduce the sampling points from 72 to 11. The rationale behind this specific number will be explained below. 

\subsubsection{Depth-Guided Sampling:} 
To mitigate border artifacts, we begin with a $256^2$ resolution depth image denoted as $D_{origin}$. Our approach involves aggregating the depths of neighboring pixels for each pixel. The formulation is as follows:
\begin{equation}
    D(x,y,i) = \left\{
    \begin{aligned}
    &D_{origin}(x+\delta_x^i,y+\delta_y^i)& &,\quad i\in \{1,...,9\} \\
    &D_{origin}(x,y)+\tau& &,\quad i\in \{10\} \\
    &D_{origin}(x,y)-\tau& &,\quad i\in \{11\}, \\
    \end{aligned}
    \right .
\end{equation}
\begin{equation}
    \delta_x^i = (i-1)/3 -1,\ \delta_y^i = (i-1)\%3-1
\end{equation}
Here, $D(x,y,i)$ represents the $i_{th}$ depth value for pixel $(x,y)$. For $i\in \{1,2,...,9\}$, $D(x,y,i)$ contains the depths of neighboring pixels. In other cases it contains depths sampled around $D_{origin}(x,y)$. To enhance the resolution of $D$, we first sort it for each pixel and then upsample it to $1024^2$ resolution. 

\subsubsection{Semantic-Guided Sampling:} In Stage 2, we address the computational cost associated with having $K$ local generators. Since our goal is to increase the resolution of the synthesized image while preserving its structure, we focus on the most important semantic part for each pixel. Specifically, we upsample the semantic mask synthesized by the generator (as described in Eq. \ref{eq:color}) to $1024^2$ resolution. By querying the weights of different semantic parts for each pixel, we can mask out those parts whose weights fall below the threshold $\delta$. 

\subsection{Training}

\subsubsection{Loss Function:}In stage 1, a low-resolution discriminator $D_{256}$, which comprises image discriminator $D_{image}$, semantic discriminator $D_{semantic}$, normal discriminator $D_{normal}$ and face discriminator $D_{face}$, is employed to train $K$ local generators. In stage 2, we freeze the $K$ generators and focus on training the 3D-aware super-resolution module, only one high-resolution image discriminator $D_{1024}$ is used in this stage. To ensure consistency between $I_{256}$ and $I_{1024}$, we introduce an upsample loss term: $\mathcal L_{upsample} = \Vert{\rm Downsample}(I_{1024})-I_{256}\Vert$. The loss functions for both stages are as follows:
\begin{equation}
    \mathcal L_{1} = \mathcal L_{256} + \mathcal L_{AG3D},\ 
    \mathcal L_{2} = \mathcal L_{1024} + \mathcal L_{upsample} + \mathcal L_{AG3D},
\end{equation}
where $\mathcal L_{256}$ and $\mathcal L_{1024}$ represent GAN loss for $D_{256}$ and $D_{1024}$ respectively. $\mathcal L_{AG3D}$ is a loss function borrowed from AG3D \cite{dong2023ag3d}.

\subsubsection{Implementation Details: }
The models are trained on 4 NVIDIA A40 GPUs for 9 days. The training in stage 1 takes 6 days, and stage 2 takes 3 days. Additionally, all the experiments mentioned in the paper were also conducted using the A40. During the training in stage 1, the normal discriminator is used only for the last 3 days of training. 

%% file: paper/experiments.tex
\section{Experiments}
\label{exp}

\subsubsection{Datasets.}
The DeepFashion dataset \cite{liu2016deepfashion} comprises 12,701 pairs of human images and corresponding semantic masks. For our training and evaluation, we utilize 8,037 pairs from this dataset. During training, we leverage a pre-trained model \cite{xiu2022icon} to obtain normal maps, , and we convert the semantic masks from 24 categories into 6 simplified categories. For instance, dresses and rompers are grouped under the broader category of "tops". Additionally, the SMPL \cite{bogo2016keep} parameters for each image are provided by AG3D \cite{dong2023ag3d}. All experiments in this paper are conducted using the DeepFashion dataset, and all models are trained on this dataset. 

\subsubsection{Metrics.}
We evaluate our method using two key metrics: Frechet Inception Distance (FID) \cite{heusel2017gans} and Kernel Inception Distance (KID) \cite{binkowski2018demystifying}. These metrics assess the diversity and quality of synthesized images by measuring their similarity to real images. Given that different methods yield relatively random results, all FID and KID scores reported in this paper are based on 50,000 synthesized images to ensure fair comparison.

\subsubsection{Baselines.}
To demonstrate the superiority of our method, we compare against several baselines: AG3D \cite{dong2023ag3d} is a SOTA method for 3d-aware human image synthesis. EVA3D \cite{hong2022eva3d} and GSM \cite{abdal2023gaussian} achieve $512^2$ resolution image synthesis without a super-resolution module. AttriHuman-3D \cite{yang2023attrihuman} enables semantic-aware generation and VeRi3D \cite{chen2023veri3d} allows local editing of synthesized images. Other works \cite{zhang2022avatargen,jiang2023humangen,yang20223dhumangan,xu2023xagen} have not released their training code or pre-trained model, so we do not include them in our comparison. 

\subsection{Comparison}
\label{comparison} 
\begin{figure}[t]
  \centering
    \begin{minipage}[t]{\columnwidth}
    \makeatletter\def\@captype{table}
    \tabcolsep=0.24cm
    \centering
    \begin{tabular}{l c c c c c c c}
    \toprule
    Method & A & B & C & D &Resolution & FID$\downarrow$ & 1000$\times$KID$\downarrow$\\
    \midrule
    AG3D & \ding{56} & \ding{56} & \ding{56} & \ding{56} & $\ \,512^\star$ & 11.33 & 5.75\\
    EVA3D & \ding{56} & \ding{56} & \ding{56} & \ding{56} & 512  & 15.89 & 9.25\\
    GSM & \ding{52} & \ding{56} & \ding{56} & \ding{56} & 512 & $\ \,15.78^\divideontimes$ & / \\
    VeRi3D & \ding{52} & \ding{56} & \ding{56} & \ding{56} & 512  & $\ \,21.4^\divideontimes$ & /\\
    AttriHuman-3D & \ding{52} & \ding{52} & \ding{56} & \ding{56} & $\ \,512^\star$  & $\ \,16.85^\divideontimes$ & / \\
    \midrule
    SemanticHuman-HD & \ding{52} & \ding{52} & \ding{52} & \ding{52} & \makecell{512 \\ 1024}  & \makecell{10.04 \\ \pmb{8.70} } & \makecell{5.02 \\ \pmb{4.04}}\\
    \bottomrule
    \end{tabular}
    \caption{Quantitative comparisons. $^\star$ denotes the use of a super-resolution module that is not 3D-aware. Some results are marked with $^\divideontimes$, indicating that these results are quoted from other papers because the authors did not release their training code or pre-trained model. A: Local editing. B: Semantic-aware synthesis. C: Semantic disentangled synthesis. D: 3D Garment generation. }
    \label{tab:fid}
  \end{minipage}
  \begin{minipage}[t]{\columnwidth} 
      \centering
      \includegraphics[width=\columnwidth]{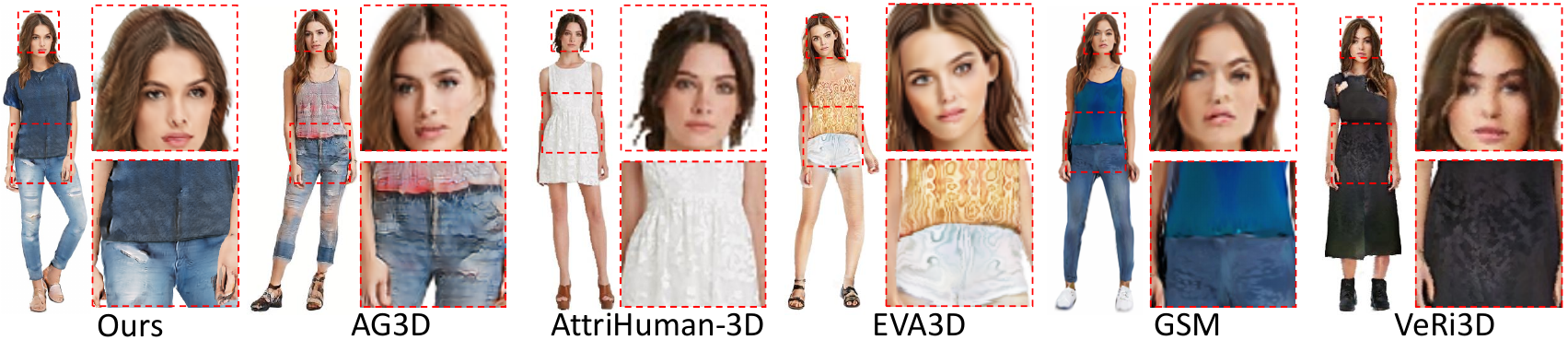}
      \caption{Qualitative comparison. To better assess the detailed quality of the generated results, we zoom in on the face and clothing areas in the synthesized images. Notably, the image synthesized by our method is at $1024^2$ resolution, whereas the results from other methods are only at $512^2$ resolution. }
      \label{fig:1024}
  \end{minipage}
\end{figure}
\subsubsection{Human Image synthesis.}
Table \ref{tab:fid} presents quantitative comparisons between our method and several SOTA methods. Our method consistently produces superior image quality at both $512^2$ and $1024^2$ resolutions. While AG3D \cite{dong2023ag3d} ranks as the second-best method in terms of image quality, it relies on a 2D super-resolution module, which compromises 3D consistency. EVA3D \cite{hong2022eva3d}, capable of synthesizing $512^2$ resolution images without a super-resolution module, unfortunately suffers from circular artifacts due to its network design. On the other hand, GSM \cite{abdal2023gaussian} and VeRi3D \cite{chen2023veri3d} allows local editing, but their lack of semantic awareness results in artifacts. AttriHuman-3D \cite{yang2023attrihuman}, while enabling semantic-aware synthesis, they entangles different semantic parts during generation. In summary, our method uniquely achieves a comprehensive set of capabilities: local editing, semantic disentangled synthesis, and 3D garment generation. Moreover, leveraging our 3D-aware super-resolution module, we stand out as the sole method capable of synthesizing $1024^2$ resolution images, as demonstrated qualitatively in Fig. \ref{fig:1024}. 

\subsubsection{Local Editing.}
\begin{figure}[t]
  \centering
  \begin{minipage}[t]{\columnwidth} 
      \centering
      \includegraphics[width=\columnwidth]{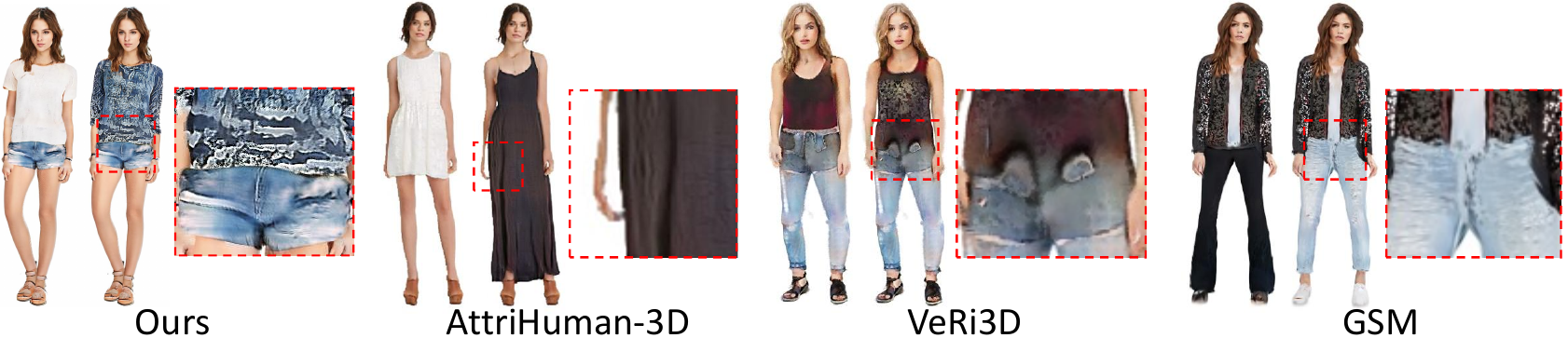}
      \caption{Comparison for local editing. For each edited image, we zoom in on key areas to demonstrate the editing capabilities. }
      \label{fig:edit}
  \end{minipage}
  \begin{minipage}[t]{\columnwidth}
      \centering
      \includegraphics[width=0.9\columnwidth]{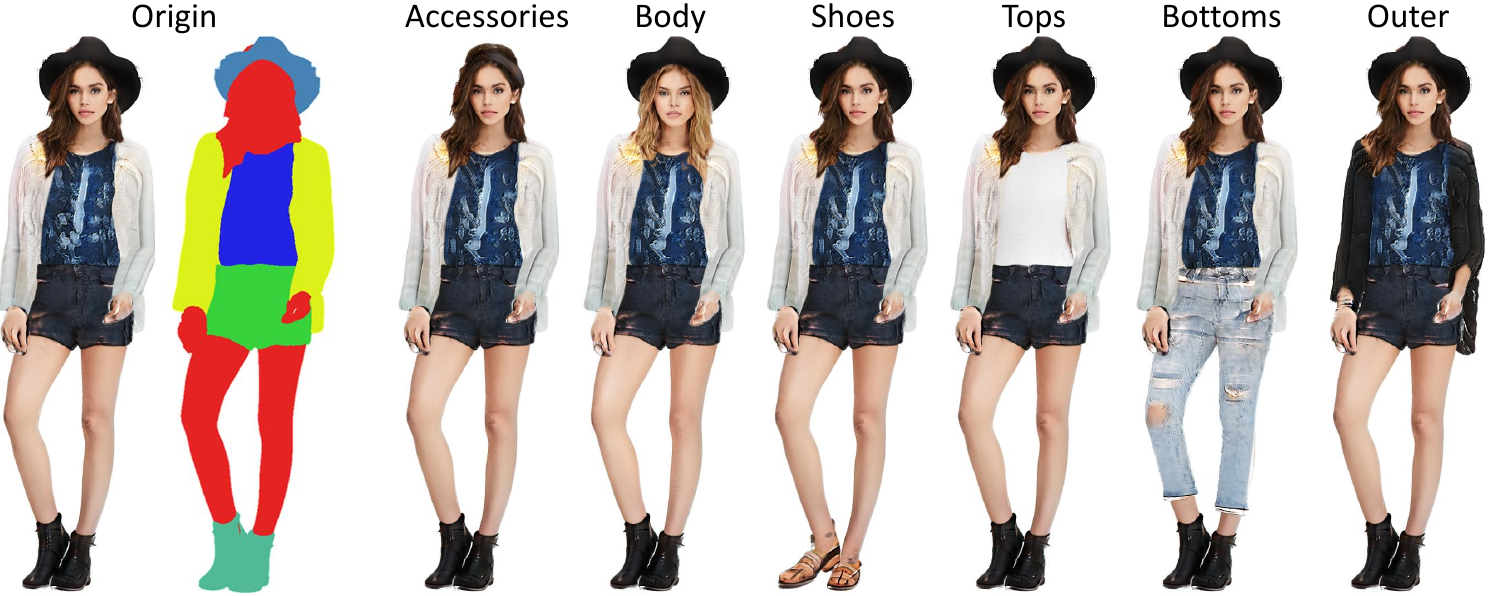}
      \caption{Semantic disentangled image synthesis. By modifying the latent code of a specified semantic part, we can alter that specified part in the synthesized image. }
      \label{fig:change}
  \end{minipage}
\end{figure}

Both AttriHuman-3D \cite{yang2023attrihuman} and our method exhibit semantic awareness. However, AttriHuman-3D employs a single generator to generate tri-plane representations corresponding to different semantic parts. Unfortunately, this design entangles different semantic parts. Consequently, when editing a specific semantic part, it may not seamlessly match other regions. GSM \cite{abdal2023gaussian} and VeRi3D \cite{chen2023veri3d} allows local editing by manipulating features of specified vertices. Although SMPL \cite{bogo2016keep} provides category labels for these vertices, the resulting edits lack semantic awareness, leading to suboptimal image quality. A comparison of the editing capabilities across different methods is illustrated in Fig. \ref{fig:edit}. For additional evidence of semantic disentangled synthesis, refer to Fig. \ref{fig:change}. 

\subsubsection{Computational Efficiency.}
Our proposed super-resolution module significantly reduces the number of required sampling points, thereby minimizing computational costs. Comparative experiments on computational resources, specifically GPU memory usage during training, are detailed in Table \ref{tab:cost}. The results clearly demonstrate that our method outperforms other methods in terms of computational efficiency. As a consequence, we can successfully synthesize $1024^2$ resolution images—a feat that other methods may struggle to achieve.

\subsection{Ablation Study}
\label{ablation}

\subsubsection{3D-Aware Super-Resolution Module.}
Our proposed super-resolution module enhances the resolution of synthesized images from $256^2$ to $1024^2$ while preserving 3D consistency. Quantitative results in Table \ref{tab:ablation} demonstrate that this super-resolution module significantly improve the quality of synthesized images. Refer to Fig. \ref{fig:sr} for qualitative results showcasing the effectiveness of this module.

\subsubsection{Depth Aggregation.}
Within our super-resolution module, we introduce depth aggregation to address discontinuities in depth maps. Consider the following scenario: an image depicts a hand positioned in front of a leg (as illustrated in Fig. \ref{fig:sr}). Suppose the depth value for the hand is approximately 1, while the depth value for the leg is approximately 2. If we directly upsample the depth map, the boundary between the leg and hand might yield an incorrect depth value of approximately 1.5. Clearly, the correct depth should align with either 1 or 2. To rectify this, we propose to aggregate depths from neighboring pixels. The qualitative results in Fig. \ref{fig:sr} validate the effectiveness of depth aggregation. 

\subsubsection{Upsample Loss.}
During super-resolution module training, we employ an upsample loss to ensure consistency between the original image and the image after super-resolution. Quantitative results reported in Table \ref{tab:ablation} confirm the effectiveness of this upsample loss.

\noindent \begin{minipage}[t]{\textwidth}
\begin{minipage}[t]{0.4\textwidth}
\makeatletter\def\@captype{table}
\tabcolsep=0.18cm
\begin{tabular}{l c c}
\toprule
Methods & Resolution & Mem \\
\midrule
EVA3D & 512 & 34G \\
AG3D & 512 & 21G \\
Ours & 512 & 10G \\
Ours & 1024 & 31G \\
\bottomrule
\end{tabular}
\caption{Efficiency comparisons. EVA3D \cite{hong2022eva3d} and AG3D \cite{dong2023ag3d} are unable to synthesize $1024^2$ resolution images. }
\label{tab:cost}
\end{minipage}
\hfill
\begin{minipage}[t]{0.58\textwidth}
\makeatletter\def\@captype{table}
\tabcolsep=0.18cm
\begin{tabular}{c c c c}
\toprule
Methods & Resolution & FID $\downarrow$ & 1000$\times$KID$\downarrow$ \\
\midrule
w/o SR & 256 & 13.47 & 9.13 \\
w/o DA & 1024 & 9.38 & 4.56 \\
w/o UL & 1024 & 13.52 & 8.18 \\
Ours & 1024 & \pmb{8.70} & \pmb{4.04} \\
\bottomrule
\end{tabular}
\caption{Quantitative ablation studies. The results validate the effectiveness of our proposed components. SR: Super-Resolution. DA: Depth Aggregation. UL: Upsample Loss.}
\label{tab:ablation}
\end{minipage}
\begin{minipage}[t]{\columnwidth}
\makeatletter\def\@captype{figure}
  \centering
  \includegraphics[width=\columnwidth]{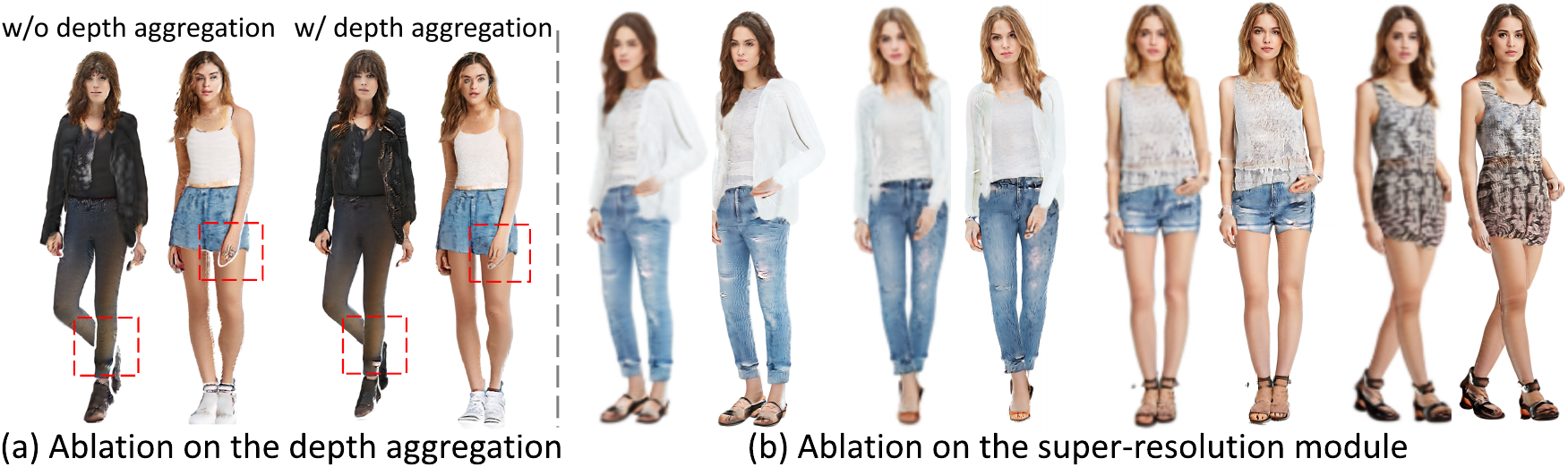}
  \caption{Qualitative ablation studies. (a) The results on the left are synthesized by the model that dose not using depth aggregation, while the images on the right are synthesized by the opposite approach. (b) For each paired set of images, the origin image is on the left, and the image after super-resolution is on the right. }
  \label{fig:sr}
\end{minipage}
\end{minipage}

\subsection{Applications}
\label{app}
Our method can achieve many interesting applications, some of which are showcased below. Additional results are provided in the supplementary material. 
\begin{figure}[t]
    \centering
    \includegraphics[width=\columnwidth]{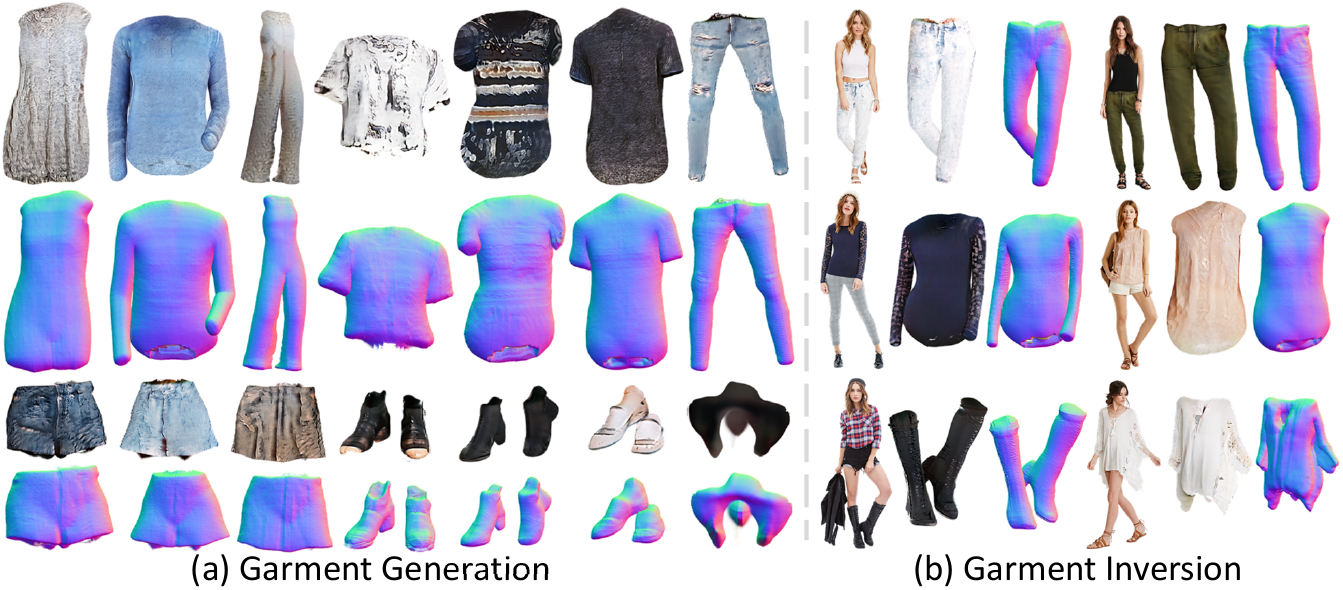}
    \caption{Garment generation. We independently generate 3D garments by setting the density of other semantic parts (except the specified part) to 0. The corresponding normal map for each synthesized image is also displayed to demonstrate the geometric quality of the results. (a) Results randomly generated by our model. (b) Results obtained from GAN inversion. }
    \label{fig:garment}
\end{figure}

\subsubsection{Semantic-Aware Virtual Try-On.}
To further demonstrate the capabilities of our method, we combine GAN inversion to achieve semantic-aware virtual try-on, as shown in Fig. \ref{fig:inv}. Notably, our method can disentangle specific garments from the results obtained by GAN inversion and even place these garments into new images. As far as we know, we are the only method that achieves this, leading to a special application: you can combine the bottoms from one image and randomly generated tops with your own image, all while controlling the pose and viewpoint of the newly synthesized image. 

\subsubsection{3D Garment Generation.}
Our method disentangles both geometry and texture, enabling 3D garment generation. Specifically, by setting the density of other semantic parts (except the specified part) to 0, we obtain the generation of specific items such as dresses, shoes, and hats. Refer to Fig. \ref{fig:garment} for results. 

\subsubsection{Out-of-Domain Image Synthesis.}
 Leveraging our disentangled synthesis, we can create out-of-domain images—think of a man wearing a dress—by manipulating the semantic latent code. Fig. \ref{fig:ood} showcases some intriguing out-of-domain image synthesis results that defy typical dataset or daily life representations.

\subsubsection{Controllable Image Synthesis.}
\begin{figure}[t]
  \centering
  \begin{minipage}[t]{\columnwidth} 
      \centering
      \includegraphics[width=\columnwidth]{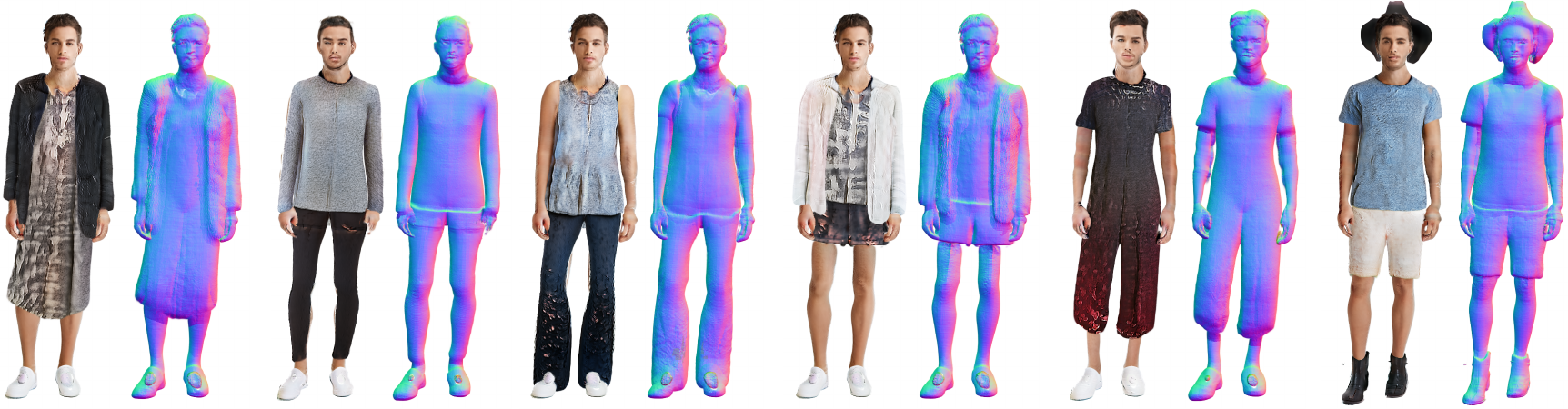}
      \caption{Out-of-domain image synthesis. To achieve out-of-domain image synthesis, we assign different semantic labels to various semantic parts, e.g., if we set the semantic label corresponding to the body as "male", and the label corresponding to the tops as "dress", we can synthesize an image of a man wearing a dress. }
      \label{fig:ood}
  \end{minipage}
  \begin{minipage}[t]{\columnwidth}
      \centering
      \includegraphics[width=\columnwidth]{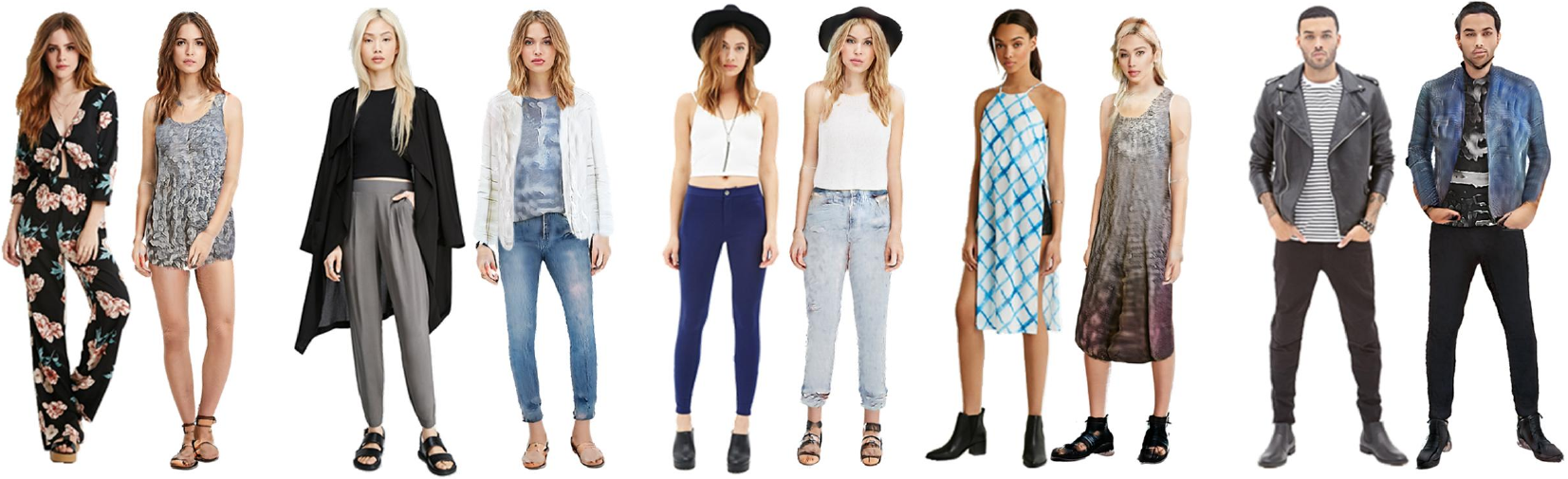}
      \caption{Conditional image synthesis. For each paired set of images, the image on the right is synthesized conditioned on the semantic label $L_s$ and human pose $P$ of the real image on the left. }
      \label{fig:cond}
  \end{minipage}
\end{figure}

The synthesis of our method is conditioned on pose $P$ and semantic label $L_s$. The results of conditional image synthesis are shown in Fig. \ref{fig:cond}. Furthermore, our method allows to control the pose of the generated 3D human and render it from various viewpoints, as shown in Fig. \ref{fig:inv}.

%% file: paper/limitation.tex
\section{Limitation}
\label{limit}
Our method faces certain limitations, which we discuss below: Dataset Constraints: The quality of synthesized results suffers when dealing with poses or viewpoints that are rarely encountered in the dataset. Unfortunately, optimizing the network alone does not fully address this issue. To overcome this limitation, we require higher-quality datasets. Challenges with 2D Supervision: While obtaining 2D human images is relatively easy, training a model only based on 2D images proves challenging when aiming for results with accurate geometries. A potential solution could involve training a model supervised by both 3D human models and 2D images, leveraging complementary information from both domains. Hand Generation Challenges: Existing methods struggle with hand generation, and achieving realistic hand deformations remains elusive. Addressing this limitation is an ongoing area of research.

%% file: paper/conclusion.tex
\section{Conclusion}
In this paper, we introduce SemanticHuman-HD, the pioneering method for achieving semantic disentangled human image synthesis. By leveraging our proposed 3D-aware super-resolution module, our method is also the first to successfully synthesize images at an impressive $1024^2$ resolution. Notably, the proposed 3D-aware super-resolution can be easily employed in NeRF-based generative models. Our experiments consistently demonstrate the superiority of our proposed method. Furthermore, we showcase a range of interesting applications, including 3D garment generation, semantic-aware virtual try-on, controllable image synthesis, garment-level image editing and out-of-domain image synthesis.  Looking ahead, we will consider addressing the limitations mentioned in Section \ref{limit}: dataset constraints, challenges with 2D supervision, and hand generation challenges. Specifically, we are particularly focused on overcoming the challenges related to 2D supervision. A generative model capable of achieving high-quality geometric details in 3D human generation is highly needed in the areas of virtual reality, video games, and beyond.

%% file: paper/supp.tex
\section{Supplementary Material}
\label{supp}
In this \textbf{Supplementary Material}, we first introduce the implementation details of our proposed SemanticHuman-HD. Furthermore, we present additional results related to SemanticHuman-HD. For video demonstrations, including animations of generated humans and 3D-aware human image synthesis, please see the \textbf{Supplementary Video}. 

\subsection{Implementation Details}
To assist others in reproducing our work or specific components, we provide implementation details of SemanticHuman-HD, including network architecture, dataset processing, training strategies an other relevant information. 

\subsubsection{Local Generator.}
For each local generator in SemanticHuman-HD, we utilize the same generator architecture as employed in AG3D \cite{dong2023ag3d}. However, to reduce the computational cost, we halve the number of channels in the network. Remarkably, this adjustment does not compromise the performance of model, as each local generator focuses solely on generating its corresponding semantic part. During the training, different local generators share the same semantic latent code as a condition, ensuring consistency across different semantic parts. In inference, we can feed different semantic latent codes to different local generators, enabling interesting applications such as semantic-aware interpolation and out-of-domain image synthesis.

\subsubsection{Discriminators.}
Throughout the training of stage 1 and stage 2, we employ multiple discriminators, including image discriminators, normal discriminators, semantic discriminators and face discriminators, to supervise the training. These discriminators share a common network architecture, only differing in channels and resolutions of input. Particularly, we opt not to use paired input of images and semantic masks. Our experiments reveals that employing two independent discriminators—one for images and another for semantic masks—enhances training stability. To facilitate conditional image synthesis, the discriminators also take human pose $P$ and semantic label $L_s$ as input. Additionally, we amplify the R1 penalty for the semantic discriminator by a factor of 10 compared to other discriminators, as we observed that introducing the semantic discriminator made the model more prone to collapse. 

\subsubsection{3D-Aware Super-Resolution Module.}
Our proposed super-resolution module leverages a feature super-resolution component to obtain high-resolution tri-plane representations. This module shares the same network design as the local generators and takes as input the low-resolution tri-plane representations generated during stage 1. Additionally, it conditions on the same semantic latent code used in stage 1. In the depth-guided sampling, we upsample the depth maps using a 2D FIR filter with coefficients [1, 3, 3, 1], which aligns with the super-resolution of tri-plane representations. Similarly, in the semantic-guided sampling, we first upsample the semantic masks and subsequently mask out those semantic parts whose values fall below the threshold $\delta$. The chosen value for $\delta$ is 0.0005, ensuring that valid semantic parts are not inadvertently masked out. Notably, even with this small $\delta$, the semantic-guided sampling can still exclude other semantic parts besides one specific part for most pixels, providing evidence of the semantic disentanglement achieved by our method.

\subsubsection{GAN Inversion.}
In GAN inversion, our goal is to derive a latent code that can be mapped to a generated result similar to a given target result. This process involves optimizing the latent code by minimizing the difference between the target results and the generated results. As in SemanticStyleGAN \cite{shi2022semanticstylegan}, we employ LPIPS \cite{zhang2018unreasonable} and L1 loss between target and generated results to supervise the optimization of latent code. The target results encompass human images, semantic masks and face images. Notably, the face images are cropped from human images based on their SMPL \cite{bogo2016keep} parameters. 

\subsubsection{Semantic Label.}
The semantic label $L_s$ serves as an indicator for gender and whether the image contains specific types of garments. It is represented as a 21-bit vector, where each value can be either 0 or 1. Apart from gender, the remaining 20 bits correspond to the following garment types: top, outer, skirt, dress, pants, leggings, headwear, eyeglass, neckwear, belt, footwear, bag, ring, wrist wearing, socks, gloves, necklace, rompers, earrings and tie. 

\subsubsection{Dataset Processing.}
The DeepFashion dataset provides images and semantic masks. The original masks contain 24 categories, and we simplified them into 6 broader categories: "Body" covers face, hair, and skin; "Tops" covers tops, dresses, and rompers; "Outer" covers outer; "Bottoms" covers skirts, pants, and leggings; "Shoes" covers socks and shoes; "Accessories" covers everything else.

\subsection{Additional Results}
In this section, we present additional results. Notably, some of these results are not attainable by existing methods \cite{dong2023ag3d,hong2022eva3d,abdal2023gaussian,chen2023veri3d,yang2023attrihuman}, such as semantic-aware interpolation and 3D garment interpolation. While view control and pose control are common applications, we specially apply them to the 3D garments disentangled from 3D humans. These novel capabilities demonstrate the versatility and effectiveness of our proposed SemanticHuman-HD.

\subsubsection{3D Human Interpolation.}
\begin{figure}[t]
    \centering
    \includegraphics[width=\columnwidth]{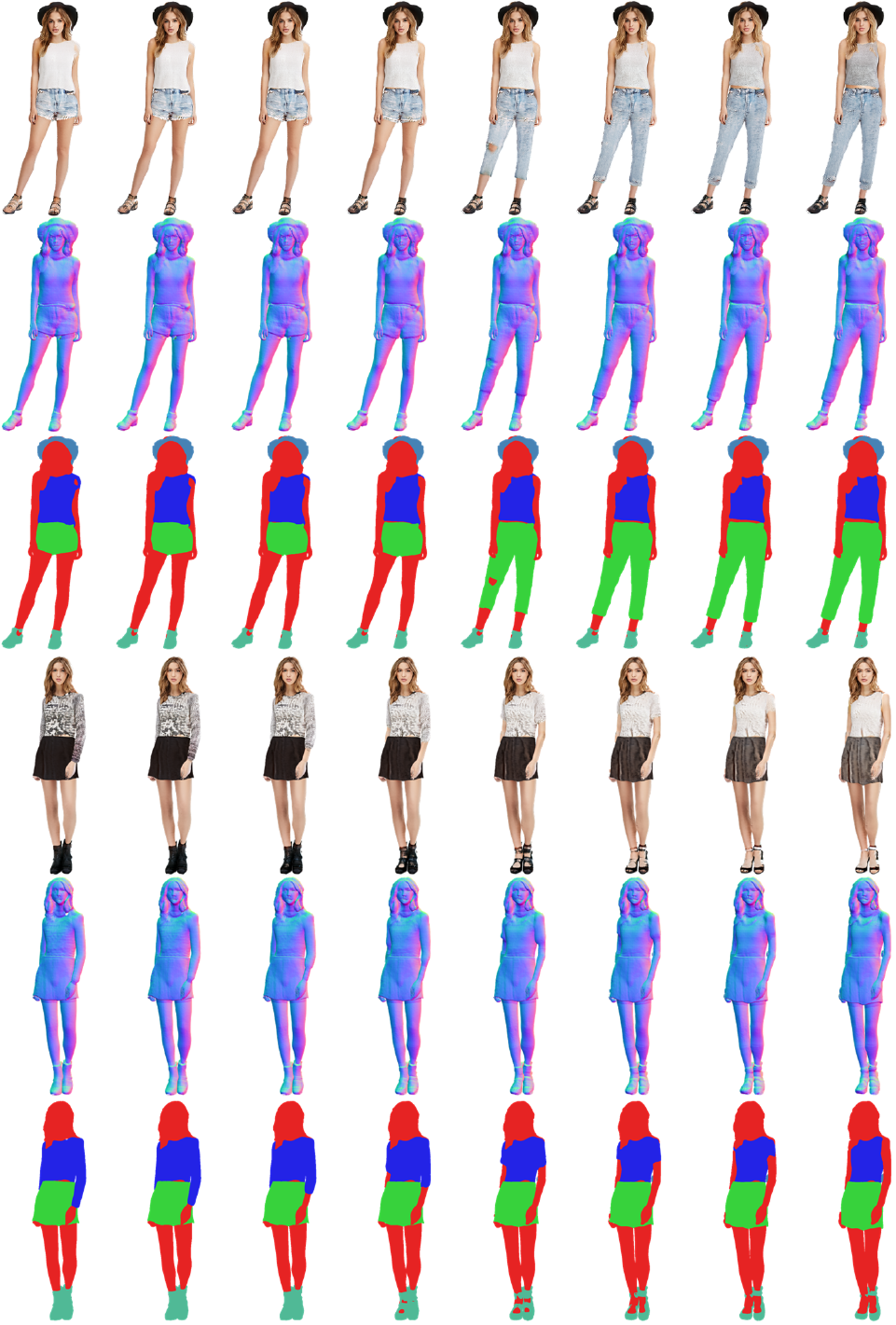}
    \caption{3D human interpolation. Each interpolation result includes an image, a normal map and a semantic mask.}
    \label{fig:interp_all}
\end{figure}

Given that each semantic latent code corresponds to a generated 3D human, interpolating between two latent codes allows for smooth transitions between two distinct 3D humans. Fig. \ref{fig:interp_all} illustrates the results of such interpolations, accompanied by corresponding semantic masks and normal maps. 

\subsubsection{Semantic-Aware Interpolation.}
\begin{figure}[t]
    \centering
    \includegraphics[width=\columnwidth]{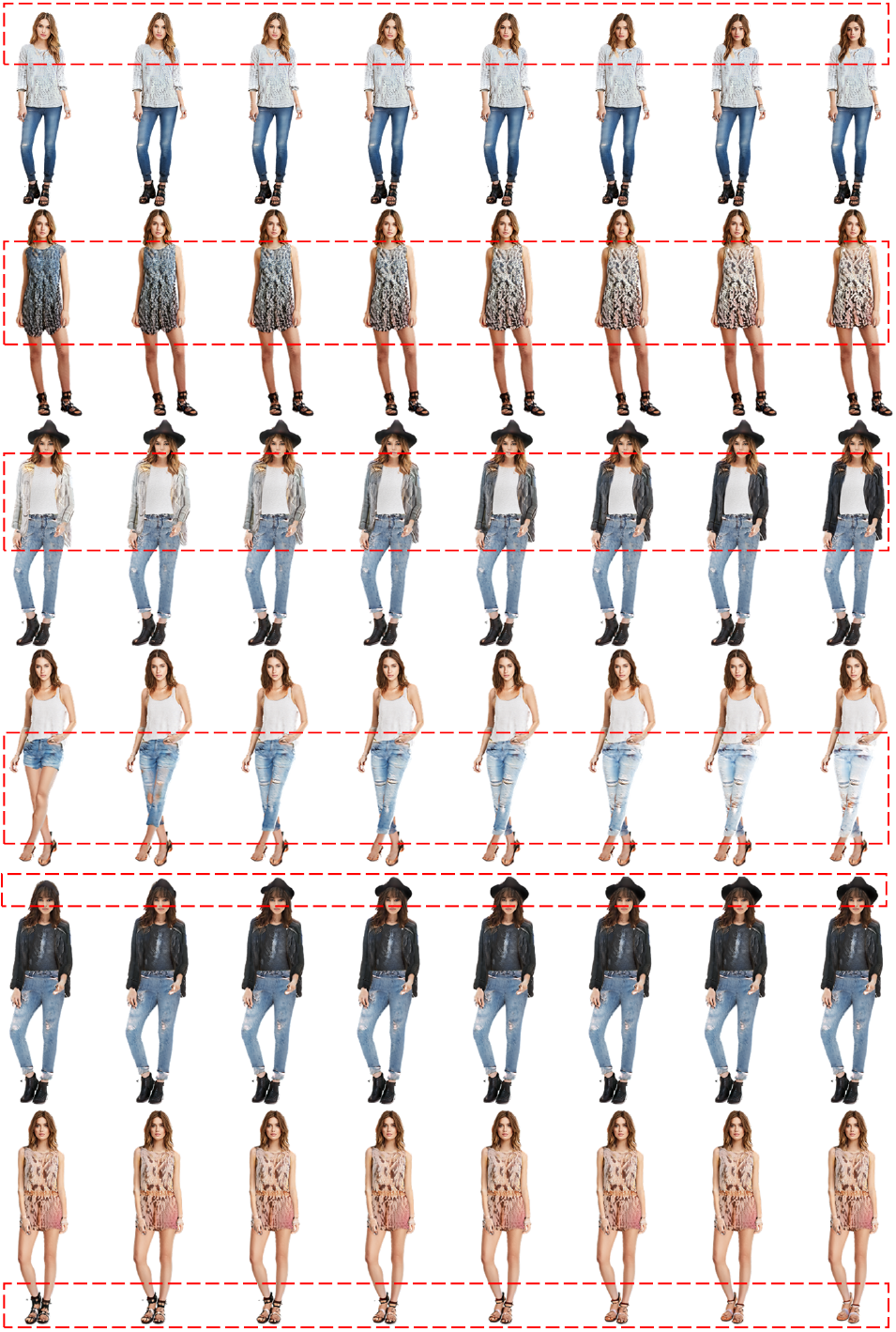}
    \caption{Semantic-aware interpolation. Red dashed rectangles on the images indicate chosen semantic parts during the interpolation.}
    \label{fig:interp_part}
\end{figure}

Our method enables semantic-aware interpolation of 3D humans. This means we can interpolate between specific semantic parts of two generated 3D human models. For instance, it allows us to smoothly alter a specific semantic part while keeping other parts unchanged. Please refer to Fig. \ref{fig:interp_part} for visual examples. The semantic-aware interpolation is achieved by interpolating between two semantic latent codes, with the interpolation confined to the codes corresponding to specific semantic parts.

\subsubsection{3D Garment Interpolation.}
\begin{figure}[t]
    \centering
    \includegraphics[width=\columnwidth]{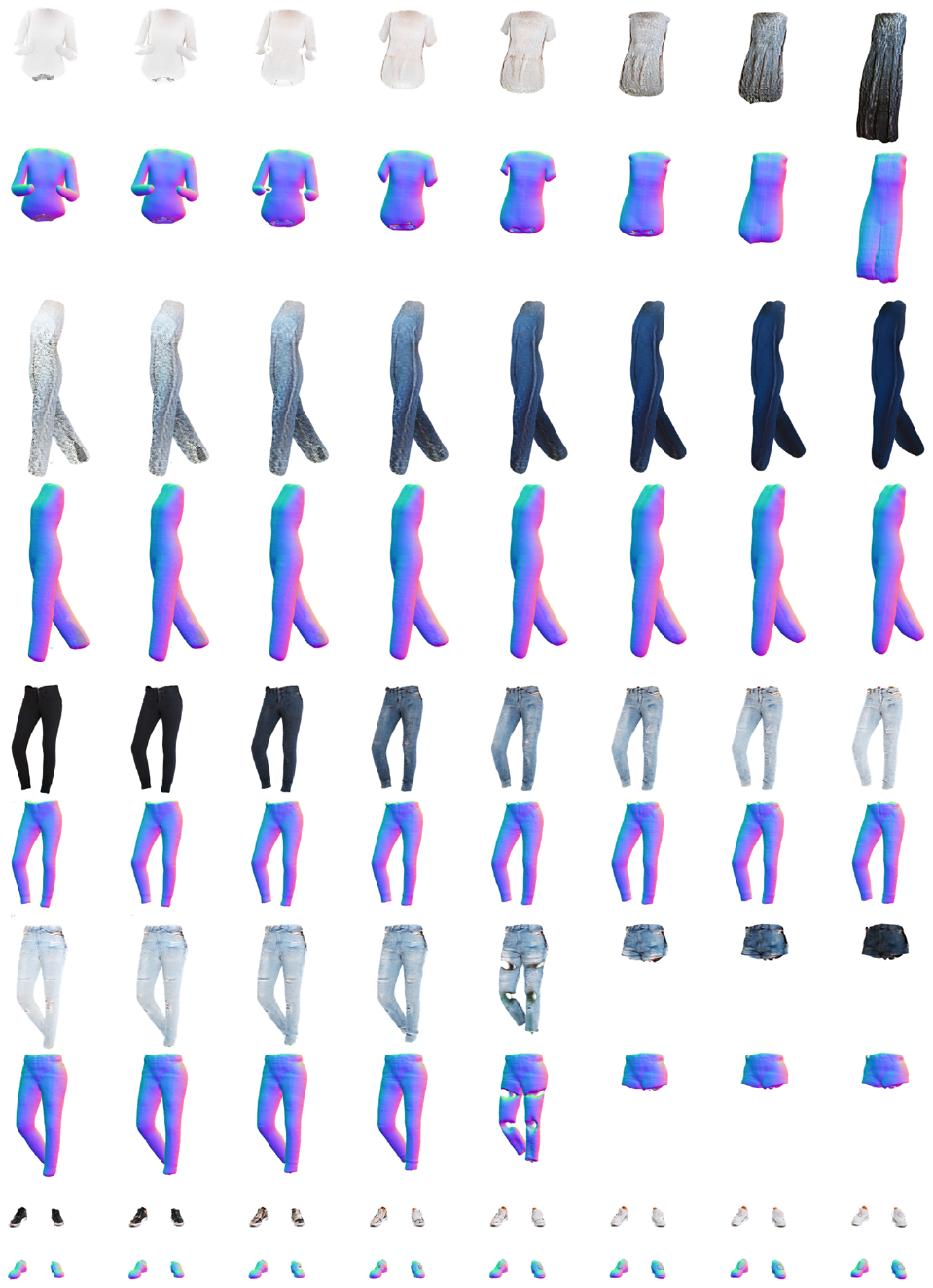}
    \caption{3D garment interpolation, including images and normal maps. For a closer view, please zoom in to see the details.}
    \label{fig:interp_garment}
\end{figure}

The independence of our generation process extends to 3D garment generation and interpolation. By setting the densities of other semantic parts (except the specific garment) to zero, we achieve 3D garment generation. Fig. \ref{fig:interp_garment} showcases the results of interpolating between two 3D garments. Additionally, we provide normal maps during the interpolation process to visualize the geometric transitions.

\subsubsection{View Control.}
\begin{figure}[t]
    \centering
    \includegraphics[width=.9\columnwidth]{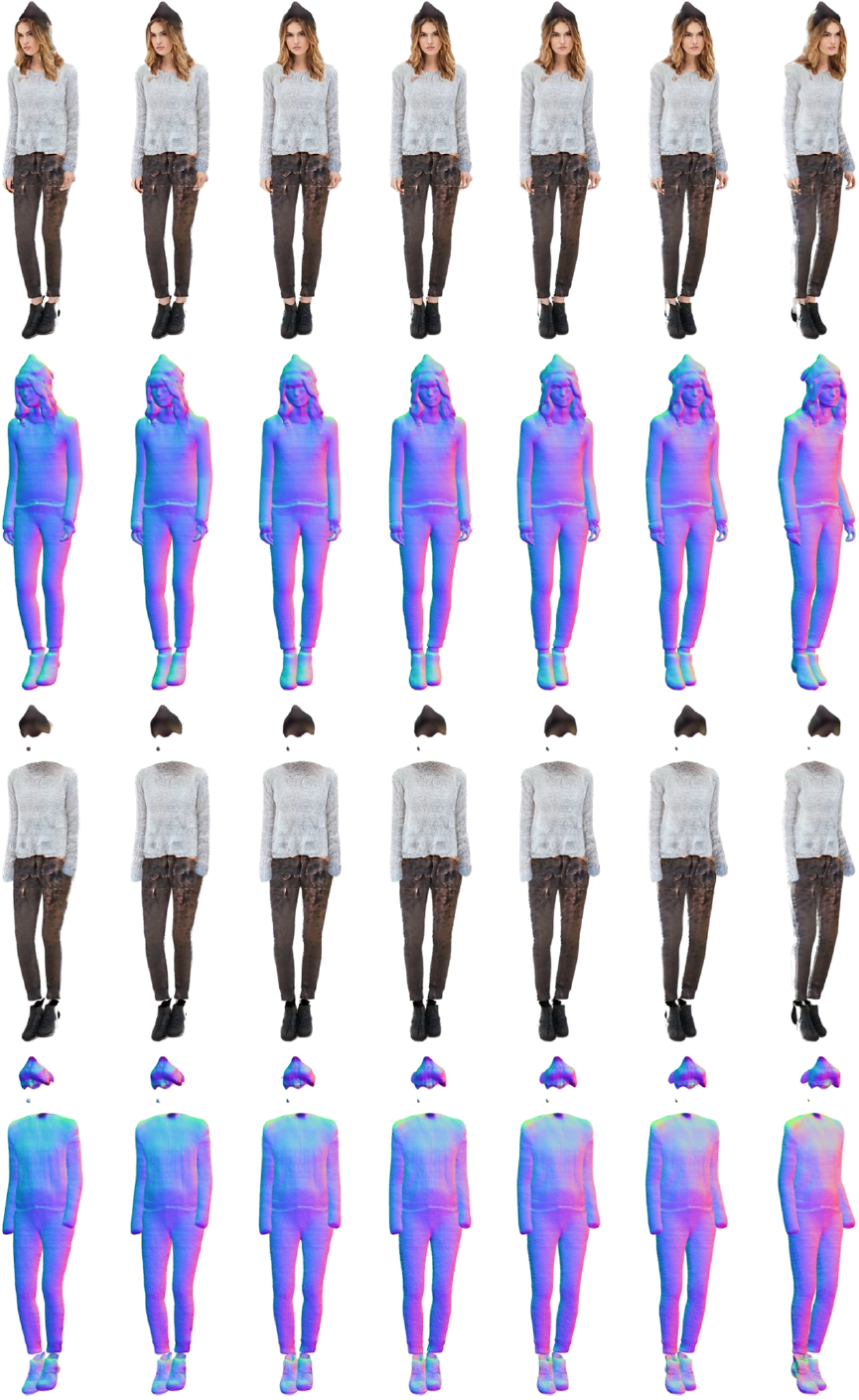}
    \caption{View control. The images in the last two rows depict 3D garments disentangled from 3D humans shown in the first two rows. }
    \label{fig:view}
\end{figure}

To demonstrate the 3D-consistent generation capability of our method, we render 3D humans and 3D garments from different viewpoints. Specifically, we disentangle the 3D garments from the generated 3D humans by setting the densities of the “body” to 0. The resulting renderings are shown in Fig. \ref{fig:view}. Notably, our proposed 3D-aware super-resolution module enables high-resolution image synthesis without compromising 3D consistency. In other words, the images rendered from different viewpoints maintain coherence. In contrast, works \cite{yang2023attrihuman,dong2023ag3d} using a 2D super-resolution module do not achieve full 3D-consistency. For a clear demonstration of this point, please refer to the videos provided by AG3D \cite{dong2023ag3d}. 

\subsubsection{Pose Control.}
\begin{figure}[t]
    \centering
    \includegraphics[width=.9\columnwidth]{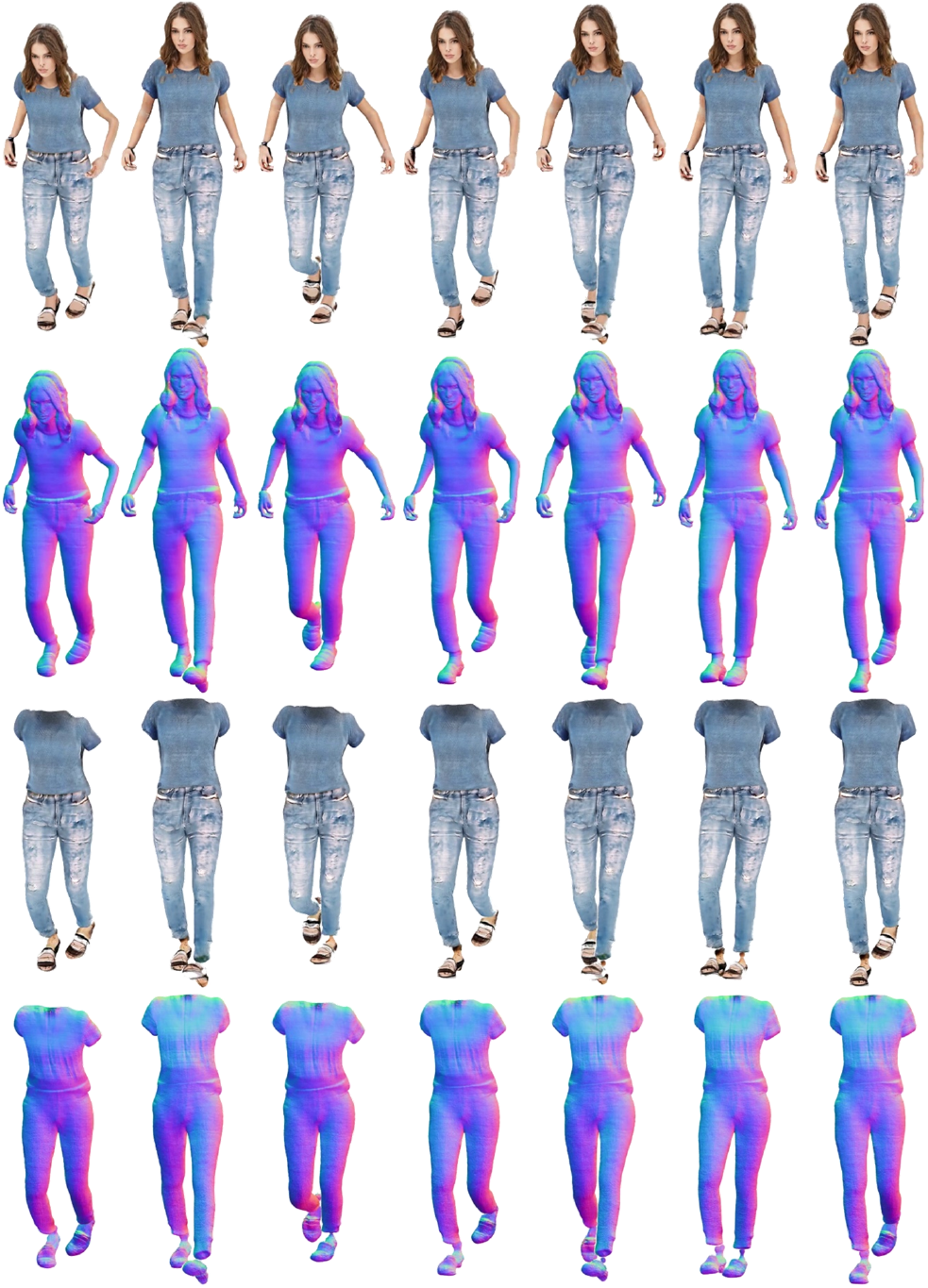}
    \caption{Pose control. The images in the last two rows depict 3D garments disentangled from 3D humans shown in the first two rows. }
    \label{fig:pose}
\end{figure}

Our method leverages the deformer to enable pose control for both generated 3D humans and 3D garments. In practical terms, this means that given a sequence of human poses, our method can animate the generated results—making them run, walk, and perform other actions. This flexibility makes our approach suitable for various downstream applications, such as virtual reality and video games. Similar to the view control, the images synthesized in different poses remain consistent, thanks to our proposed 3D-aware super-resolution module. The results are shown in Fig. \ref{fig:pose}. For a comprehensive view of the animations, please refer to the \textbf{Supplementary Video}.

